\pdfoutput=1

\documentclass[11pt]{article}

\usepackage{acl}

\usepackage{times}
\usepackage{latexsym}

\usepackage[T1]{fontenc}

\usepackage[utf8]{inputenc}

\usepackage{microtype}

\usepackage{inconsolata}
\usepackage{times}
\usepackage{latexsym}
\usepackage{makecell}
\usepackage{float}
\usepackage{amssymb}
\usepackage{multirow}
\usepackage{diagbox}
\usepackage{bm}
\usepackage{galois}
\usepackage{mathrsfs}
\usepackage{appendix}
\usepackage{verbatim}
\usepackage{caption}
\usepackage{soul}
\usepackage{graphicx}
\usepackage{amsmath}
\usepackage{amsthm}
\usepackage{helvet}
\usepackage{courier}
\usepackage{color}
\usepackage{booktabs}
\usepackage{color, colortbl}
\definecolor{Gray}{gray}{0.9}
\definecolor{LightCyan}{rgb}{0.88,1,1}
\def\argmax{\mathop{\rm argmax}}

\usepackage{algorithm}
\usepackage{subcaption}
\usepackage{colortbl}
\usepackage{array}
\usepackage{pifont}
\usepackage{CJKutf8}
\usepackage{algpseudocode}

\title{Improving Cross-lingual Representation for Semantic Retrieval with Code-switching}

\author{
Mieradilijiang Maimaiti$^{1,2,3,4}$\thanks{\ \ Equal contribution}\;\thanks{\ \ Corresponding author: Mieradilijiang Maimaiti}\;, Yuanhang Zheng$^{5*}$, Ji Zhang$^6$, \\
\textbf{Yue Zhang$^7$, Wenpei Luo$^8$, Kaiyu Huang$^9$}
\\
$^1$School of Computer Science and Technology, Xinjiang University \\
$^2$Xinjiang Laboratory of Multi-Language Information Technology, Xinjiang University \\
$^3$Xinjiang Multilingual Information Technology Research Center, Xinjiang University \\
$^4$Joint International Research Laboratory of Silk Road Multilingual Cognitive Computing,\\Xinjiang University \\
$^5$Department of Computer Science and Technology, Tsinghua University, Beijing, China \\
$^6$Alibaba DAMO Academy \\
$^7$Department of Computer Science and Technology, Westlake University, Hangzhou, China \\
$^8$Department of Computer Science and Technology, Dalian University of Technology, Dalian \\
$^9$Beijing Key Lab of Traffic Data Analysis and Mining, Beijing Jiaotong University, China \\
\tt{miradeljan51@xju.edu.cn, zheng-yh19@mails.tsinghua.edu.cn,}\\
\tt{zj122146@alibaba-inc.com, zhangyue@westlake.edu.cn,}\\
\tt{22109239@mail.dlut.edu.cn, kyhuang@bjtu.edu.cn}
}

\begin{document}
\begin{CJK}{UTF8}{gbsn}

\maketitle
\begin{abstract}
Semantic Retrieval (SR) has become an indispensable part of the FAQ system in the task-oriented question-answering (QA) dialogue scenario. 
The demands for a cross-lingual smart-customer-service system for an e-commerce platform or some particular business conditions have been increasing recently. 
Most previous studies exploit cross-lingual pre-trained models (PTMs) for multi-lingual knowledge retrieval directly,
while some others also leverage the continual pre-training before fine-tuning PTMs on the downstream tasks.
However, no matter which schema is used, the previous work ignores to inform PTMs of some features of the downstream task, i.e. train their PTMs without providing any signals related to SR.
To this end, in this work, we propose an Alternative Cross-lingual PTM for SR via code-switching.
We are the first to utilize the code-switching approach for cross-lingual SR.
Besides, we introduce the novel code-switched continual pre-training instead of directly using the PTMs on the SR tasks. 
The experimental results show that our proposed approach consistently outperforms the previous SOTA methods on SR and semantic textual similarity (STS) tasks with three business corpora and four open datasets in 20+ languages.
\end{abstract}

\section{Introduction}
In recent years, pre-trained models (PTMs) have demonstrated success on many downstream tasks of natural language processing (NLP). 
Intuitively, PTMs such as ELMO \citep{elmo_naacl18}, GPT \citep{GPT}, GPT-2 \citep{GPT2}, GPT-3 \citep{GPT3}, BERT \citep{bert_naacl19}, and RoBERTa \citep{DBLP:journals/corr/abs-1907-11692} have achieved remarkable results by transferring knowledge learned from a large amount of unlabeled corpus to various downstream NLP tasks. 
To learn the cross-lingual representations, previous methods like multi-lingual BERT (mBERT) \citep{bert_naacl19} and XLM \citep{xlm_nips19} have extended PTMs to multiple languages.

\begin{figure}[!t]
\centering
\includegraphics[width=7.5cm,height=4.5cm]{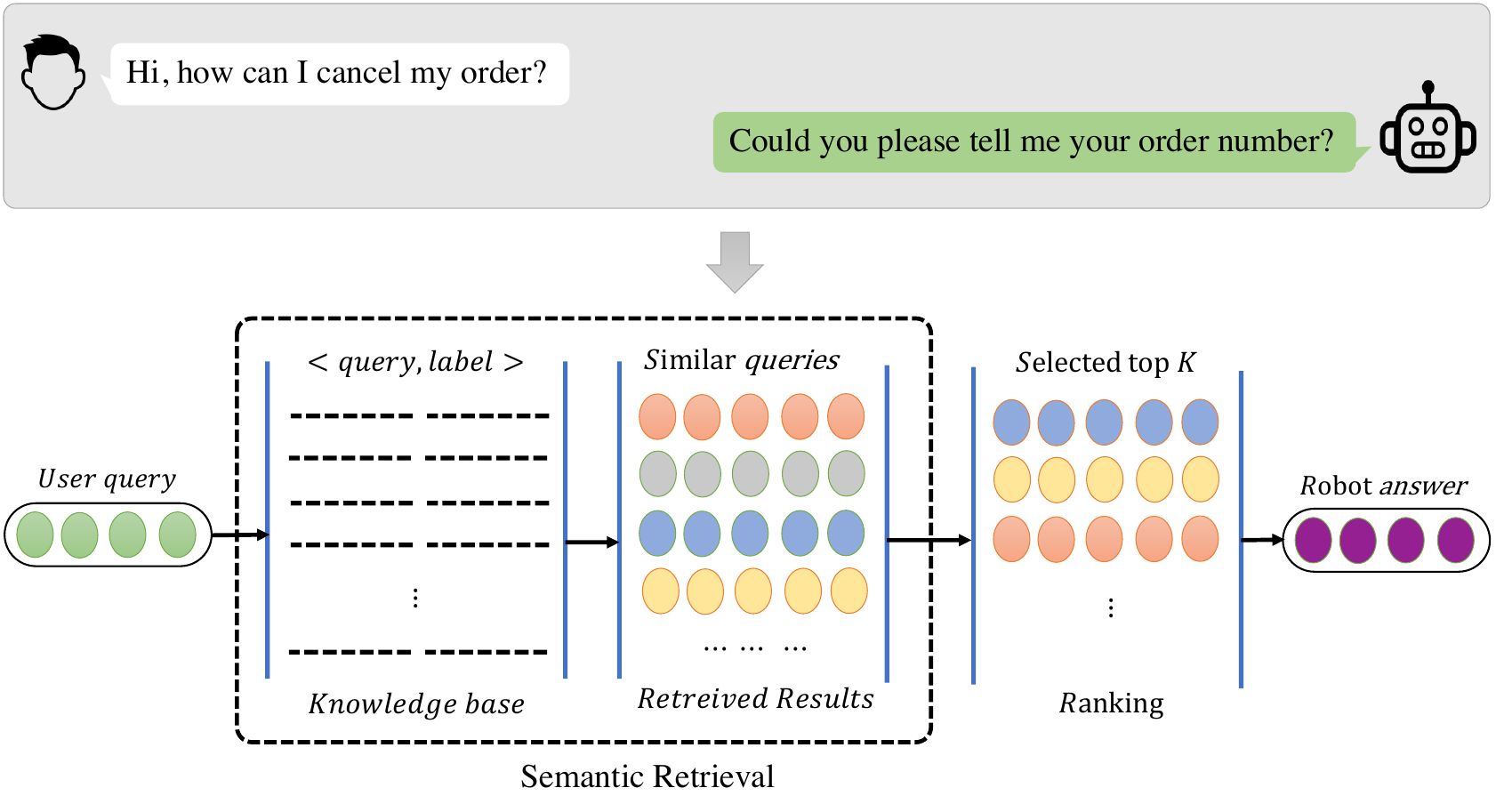}
\caption{\label{fig:lb_sen_level_sr} The brief illustration of the semantic retrieval with leveraging knowledge base for FAQ system in the task-oriented dialogue scenario.}
\end{figure}

Semantic retrieval (SR) \citep{skipthought_nips15} has become the ubiquitous method in the FAQ system (i.e., task-oriented question-answering (QA) \citep{DBLP:conf/iclr/XiongXLTLBAO21}) which is incorporated into the smart-customer-service platform for the e-commerce scenario. 
For the cross-lingual scenario, many pre-training methods have been presented for the multi-lingual downstream tasks, such as XLM-R \citep{xlm-r_acl2020}, XNLG \citep{xnlg_aaai2020}, InfoXLM \citep{info-xlm_naacl21} and VECO \citep{veco_acl2021}. 
Intuitively, the main challenge of SR is how to accurately retrieve the corresponding sentence from the knowledge base (query-label pairs) \citep{skipthought_nips15}. 
Commonly used approaches mainly take some variants of the BERT model as a backbone and then directly fine-tune on the downstream tasks \citep{bert_naacl19,xlm_nips19,unicoder_emnlp19,universal_sen_rep2020,ernie-m_emnlp21}.

Specifically, mBERT \citep{bert_naacl19}, XLM \citep{xlm_nips19}, Unicoder \citep{unicoder_emnlp19}, CMLM \citep{universal_sen_rep2020} and ERNIE-M \citep{ernie-m_emnlp21} learn the cross-lingual sentence representation mainly using masked language modeling (MLM). For other objective functions,
MMTE \citep{MMTE_AAAI20} exploits multi-lingual machine translation, and CRISS \citep{criss_nips2020} leverages unsupervised parallel data mining. 
Some models use Siamese network architectures to adapt them to SR better.
For example, InferSent \citep{infersent_emnlp17} uses natural language inference (NLI) datasets to train the Siamese network. USE \citep{use_ijcai19}, M-USE \citep{muse_acl20} and LaBSE \citep{labse_acl22} exploit ranking loss. SimCSE \citep{SimCSE21}, InfoXLM \citep{info-xlm_naacl21} and HICTL \citep{hictl_iclr21} use contrastive learning.
However, the previous highly similar approaches almost ignore the transmission of some features of the downstream tasks to PTMs.
In other words, most methods \citep{info-xlm_naacl21,veco_acl2021} directly fine-tune the models on downstream tasks without providing any signals related to SR. 
In addition, they are mainly pre-trained on combined monolingual data where few of the sentences are code-switched.
Since the user queries often contain many code-switched sentences, it is insufficient to exploit the commonly used methods directly for the SR task in the e-commerce scenario.

In this work, as depicted in Figure \ref{fig:lb_sen_level_sr},
we aim to enhance the performance of SR for the FAQ system 
in the e-commerce scenario.
We propose a novel pre-training approach for sentence-level SR with code-switched cross-lingual data.
Our motivation comes from the ignorance of previous studies.
One of the recent studies \citep{alm2020aaai} also tries to exploit the code-switching strategy in the machine translation scenario, but no one has tried to leverage code-switching on the task of multi-lingual SR.
Furthermore, the previous methods \citep{auto-mlm22} have exploited multi-lingual PTMs on the SR task by only masking the query instead of masking the label.
They intend to use more efficient PTMs to fine-tune the SR task rather than making PTMs stronger by providing some signals.
To allow the PTMs to learn the signals directly related to downstream tasks, we present an \textbf{A}lternative \textbf{C}ross-\textbf{L}ingual PTM for semantic retrieval using code-switching, which consists of three main steps.
First, we generate code-switched data based on bilingual dictionaries.
Then, we pre-train a model on the code-switched data using a weighted sum of the alternating language modeling (ALM) loss \citep{alm2020aaai} and the similarity loss.
Finally, we fine-tune the model on the SR corpus. By providing additional training signals related to SR during the pre-training process, our proposed approach can learn better about the SR task. 
Our main contributions are as follows:
\begin{itemize}
    \item  Experiments show that our approach remarkably outperforms the SOTA methods with various evaluation metrics.
    \item Our method improves the robustness of the model for sentence-level SR on both the in-house datasets and open corpora.
    \item To the best of our knowledge, we first present alternative cross-lingual PTM for SR using code-switching in the FAQ system (e-commerce scenario).
\end{itemize}

\section{Preliminaries}

\subsection{Masked Language Modeling}

Masked language modeling (MLM) \citep{bert_naacl19} is a pre-training objective focused on learning representations of natural language sentences. When pre-training a model using MLM objectives, we let the model predict the masked words in the input sentence. Formally, we divide each sentence $\mathbf{x}$ into the masked part $\mathbf{x}_m$ and the observed part $\mathbf{x}_o$, and we train the model (which is parameterized by $\mathbf{\theta}$) to minimize
\begin{equation}
\label{eq:bert_loss}
    \mathcal{L}_{MLM}=-\log P(\mathbf{x}_m|\mathbf{x}_o;\mathbf{\theta}).
\end{equation}

When calculating Eq.\eqref{eq:bert_loss}, we assume that the model independently predicts each masked word. 
Formally, we assume that all masked words $x_i$ in the masked part $\mathbf{x}_m$ are independent conditioned on $\mathbf{x}_o$.
Thus, Eq.\eqref{eq:bert_loss} can be rewritten as
\begin{equation}
     \mathcal{L}_{MLM}=-\sum_{x_i \in \mathbf{x}_m}\log P(x_i|\mathbf{x}_o;\mathbf{\theta}).
\end{equation}

\begin{figure*}[!t]
\centering
\includegraphics[trim={1cm 3cm 0.5cm 3cm},clip,width=15.5cm,height=6.0cm]{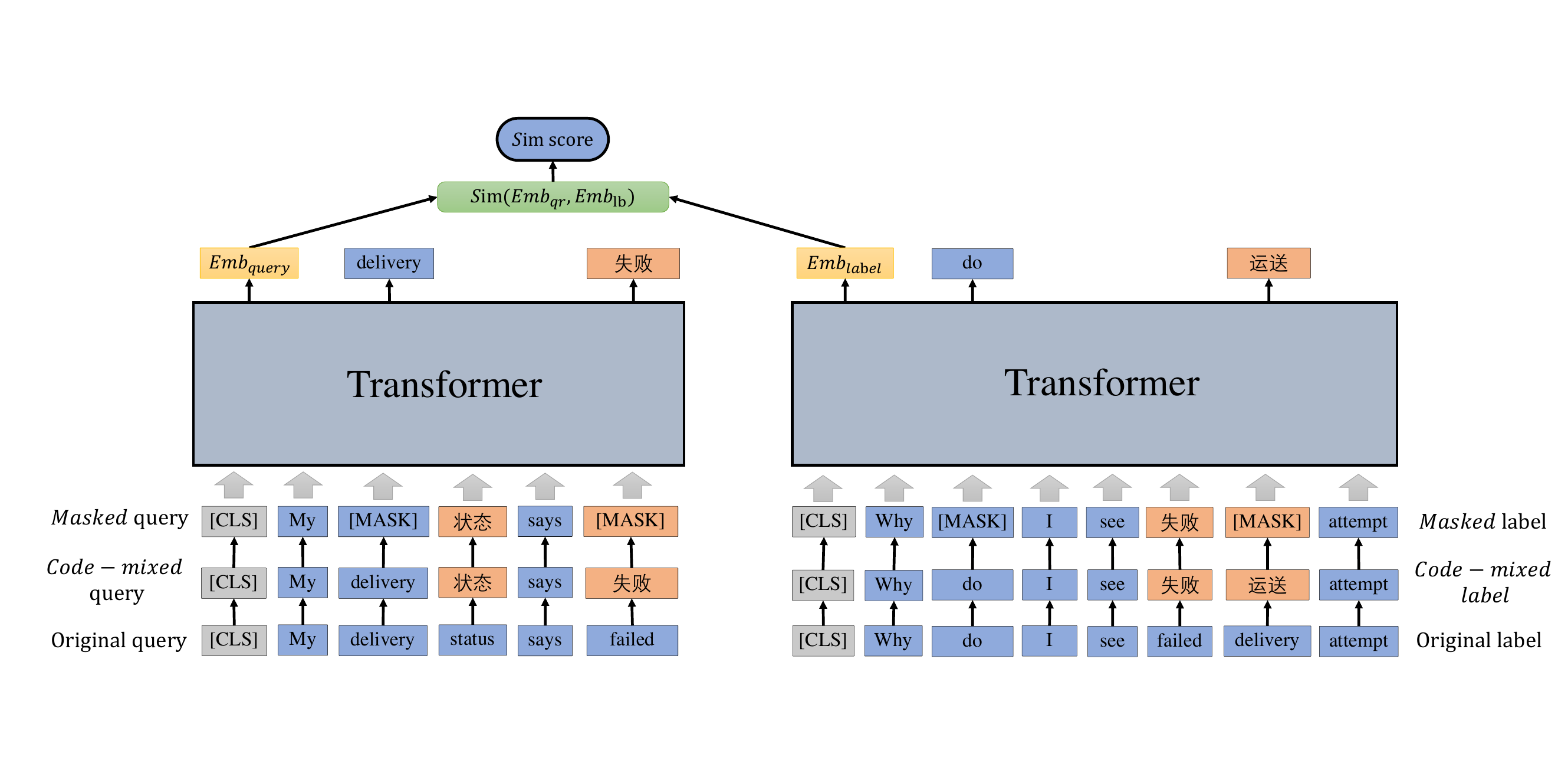}
\caption{\label{fig:model_arch} The architecture of our proposed model \textbf{A}lternative \textbf{C}ross-\textbf{L}ingual PTM for SR. The code-switched tokens for query and label are ``status"$\Rightarrow$``状态", ``failed"$\Rightarrow$``失败" and ``"delivery"$\Rightarrow$``运送" separately. The "[CLS]" symbol stands for the sentence representation of query and label. The structures used for the query and the label are the same.}
\end{figure*}

\subsection{Cross-lingual LM Pre-training}

To improve the performances of various models on the NLP tasks of different languages, cross-lingual PTMs \citep{bert_naacl19,xlm_nips19,xlm-r_acl2020} have been proposed.
Generally, cross-lingual PTMs are trained on multi-lingual corpora using the MLM objective. During the pre-training process, the corpora of low-resource languages are usually oversampled to improve the model's performance on low-resource languages.
To better align the representations of the sentences in different languages, cross-lingual PTMs may use another objective called translation language modeling (TLM), which requires the model to predict the masked words in both the source and the target sentences in a parallel sentence pair.
Formally, given a parallel sentence pair $\langle\mathbf{x},\mathbf{y}\rangle$, we randomly divide the source sentence $\mathbf{x}$ into the masked part $\mathbf{x}_m$ and the observed part $\mathbf{x}_o$, and also divide the target sentence $\mathbf{y}$ into the masked part $\mathbf{y}_m$ and the observed part $\mathbf{y}_o$. Then we minimize
\begin{equation}
\label{eq:tlm_loss}
    \mathcal{L}_{TLM}=-\log P(\mathbf{x}_m,\mathbf{y}_m|\mathbf{x}_o,\mathbf{y}_o;\mathbf{\theta}).
\end{equation}

\subsection{Semantic Retrieval}

Semantic retrieval (SR) aims to retrieve sentences similar to the query sentence in a knowledge base \citep{skipthought_nips15}. Specifically, the semantic retrieval model converts sentences into vectors, and similar sentences are retrieved based on the cosine similarity.

Formally, given a sentence $\mathbf{x}$, the model encodes $\mathbf{x}$ into a vector $\mathbf{v}_\mathbf{x}$. When we need to retrieve sentences similar to the query $\mathbf{q}$, we calculate the cosine similarity between $\mathbf{v}_\mathbf{q}$ and $\mathbf{v}_\mathbf{x}$ for each sentence $\mathbf{x}$ in the knowledge base $\mathcal{K}$:
\begin{equation}
    sim(\mathbf{q},\mathbf{x})=\frac{\mathbf{v}_\mathbf{q}\cdot\mathbf{v}_\mathbf{x}}{||\mathbf{v}_\mathbf{q}||\times||\mathbf{v}_\mathbf{x}||}.
\end{equation}

Finally, we retrieve the sentence $\mathbf{x}^{*}$ which is most similar to the query $\mathbf{q}$ in $\mathcal{K}$:
\begin{equation} \mathbf{x}^{*}=\argmax\limits_{\mathbf{x}\in\mathcal{K}}sim(\mathbf{q},\mathbf{x}).
\end{equation}

\subsection{Code-switching}

To reduce the representation gap between words of different languages in the cross-lingual PTMs. \citet{alm2020aaai} proposed
ALM, which is based on code-switching.
Specifically, given a source sentence, we construct a code-switched sentence by randomly replacing some source words with the corresponding target words.
For example, suppose the English source sentence is ``I like music'' and then we replace some words in the sentence with Chinese. If the replaced English words are ``I'' and ``music'' and their corresponding Chinese words are ``我'' and ``音乐'', respectively, then the code-switched sentence is ``我 like 音乐''.

Formally, suppose that we conduct code-switching on a source sentence $\mathbf{x}=\{x_1,x_2,\dots,x_n\}$. First, we randomly choose a subset $S$ from $\{1,2,\dots,n\}$. Then, for each element $i{\in}S$, we replace $x_i$ with its corresponding target word $y_i$ to construct the code-switched sentence $\mathbf{z}=\{z_1,z_2,\dots,z_n\}$, where
\begin{equation}
    z_i=\begin{cases}
        x_i & i{\notin}S, \\
        y_i & i{\in}S.
    \end{cases}
\end{equation}

\section{Method}    
\subsection{Alternative Cross-lingual PTM}
The main architecture of our model is shown in Figure~\ref{fig:model_arch}.
We jointly train the PTM on the code-switched data using the cross-lingual masked language model (XMLM) and the similarity loss
to address the limitation of existing PTMs that are trained without signals directly related to the downstream tasks (e.g., SR).
Contrarily, we add a similarity loss term to the pre-training objective to adjust the similarity between input ($query$) and output ($label$).
Thus, the similarity between $query$ and $label$ has been controlled by exploiting the similarity loss during the continual pre-training step.
 For the sentence-level SR task, the given knowledge is composed of a certain number of $\langle query, label \rangle$ pairs (see Figure \ref{fig:model_arch}). 
We regard $\mathbf{q}$ as a $query$ and take $\mathbf{l}$ as a corresponding $label$.
Given a query $\mathbf{q}= \{q_1,\dots,q_i,\dots,q_I\}$ and a label $\mathbf{l}=\{l_1, \dots, l_j, \dots, l_J\}$, the standard retrieval models usually formulate the sentence-level SR as a calculation of the similarity between $query$ and $label$ on the semantic space:
\begin{align}
    \mathbf{v}_\mathbf{q} &= encode(q_1,\dots,q_i,\dots,q_I), \\
    \mathbf{v}_\mathbf{l} &= encode(l_1,\dots,l_j,\dots,l_J).
\end{align}

\begin{algorithm}[!t]
\caption{Cross-lingual SR with Code-switching} \label{alg:mixture_tl}
\begin{algorithmic}[1]
\Require user query $Q_{user}=\{\mathbf{q}_{user}^{(u)}\}_{u=1}^U$,
monolingual knowledge (query-label pairs) $\mathcal{K}_{mono}=\{\langle \mathbf{q}^{(m)},\mathbf{l}^{(m)}\rangle \}_{m=1}^M$,
Bi-lingual dictionary $D_{bi}=\{\langle L_1^{(n)},L_{en}^{(n)} \rangle \}_{n=1}^N$;

\Ensure retrieved top-$k$ similar question $Q^{\prime}$;

\State Obtain code-switched knowledge $\mathcal{K}_{cmd}$ using $D_{bi}$ on $\mathcal{K}_{mono}$

\For{N knowledge-pair in $\mathcal{K}_{cmd}$} \Comment{training}
    	\State achieve the similarity $ sim(\mathbf{q},\mathbf{l})$ (Eq.\ref{eq_sim_loss})
    	\State Obtain  the code-switched $\mathcal{L}^{(\mathbf{q},\mathbf{l})}_{XMLM}$ 
    	\State Jointly optimize the total loss (Eq.\ref{eq_joint_loss})
\EndFor
\State Return retrieved top $k$ $Q^{\prime}$ from $\mathcal{K}_{cmd}$ according to $Q_{user}$

\end{algorithmic}
\end{algorithm}

We retrieve the query $\mathbf{q}$ similar to the label $\mathbf{l}$ by calculateing the cosine similarity between $\mathbf{v}_\mathbf{q}$ and $\mathbf{v}_\mathbf{l}$ for each $\langle query ,label \rangle$ pair in the knowledge base $\mathcal{K}$:
\begin{equation}
sim(\mathbf{q},\mathbf{l})=\frac{\mathbf{v}_\mathbf{q}\cdot\mathbf{v}_\mathbf{l       }}{||\mathbf{v}_\mathbf{q}||\times||\mathbf{v}_\mathbf{l}||}.
    \label{eq_sim_loss}
\end{equation}

Then, we rank the retrieved sentences according to their similarity score to recall the Top-$k$ similar asking questions that are semantically close to the original input query.
The total objective function of our proposed model consists of two parts, the XMLM and the similarity loss, which can be formulated as follows:
\begin{eqnarray}
    \mathcal{L}_{total} = \mathbf{\lambda} * \mathcal{L}^{(\mathbf{q},\mathbf{l})}_{XMLM} + \mathcal{L}^{(\mathbf{q},\mathbf{l})}_{sim},
    \label{eq_joint_loss}
\end{eqnarray}
\noindent where $\lambda > 0$ controls the weight of the XMLM.

Intuitively, since the XMLM is highly similar to the monolingual MLM, the masked token prediction task can be extended to the cross-lingual settings. Generally, the monolingual MLM loss is as follows:
\begin{eqnarray}
\mathcal{L}^{(\mathbf{q})}_{MLM} = - \sum_{q_i \in \mathbf{q}_m} \log P(q_i| \mathbf{q}_o; \mathbf{\theta}),
\end{eqnarray}

\noindent where $\mathbf{q}_m$ and $\mathbf{q}_o$ are the masked part and the observed part of the input $query$, respectively. The masked version of the input $label$ is also similar to $query$, i.e. we also mask the $label$ by using the same masking strategy of $query$.

Concretely, as shown in Algorithm \ref{alg:mixture_tl}, we merge the pairs of the $\langle query, label \rangle$ with the code-switched format, and regard it as the input of MLM. The XMLM is as follows:
\begin{equation}
\begin{aligned}
\mathcal{L}^{(\mathbf{q},\mathbf{l})}_{XMLM} = &- \sum_{q_i \in \mathbf{q}_m} \log P(q_i| \mathbf{q}_o; \mathbf{\theta}) \\
&- \sum_{l_j \in \mathbf{l}_m} \log P(l_j| \mathbf{l}_o; \mathbf{\theta}),
\end{aligned}
\end{equation}

\noindent where $\mathbf{l}_m$ and $\mathbf{l}_o$ are the masked part and the observed part of the $label$, respectvely. Besides, we provide additional training signals related to the downstream task 
during the continual pre-training process. Specifically, we expect the vectorized representation of the query $\mathbf{q}$ to be close to its corresponding label $\mathbf{l}$, but far from any incorrect label $\mathbf{l}' \neq \mathbf{l}$. To achieve this, we define the similarity loss as:
\begin{equation}
    \mathcal{L}^{(\mathbf{q},\mathbf{l})}_{sim} = -\log \frac{\exp sim(\mathbf{q},\mathbf{l})}{\exp sim(\mathbf{q},\mathbf{l})+\sum\limits_{\mathbf{l}' \in \mathcal{B}}\exp sim(\mathbf{q},\mathbf{l}')},
\end{equation}

\noindent where $\mathcal{B}$ denotes the set of all labels in a training batch other than $\mathbf{l}$.

\begin{table*}[!t]
\centering
\small
\caption{Characteristics of our business corpus. ``Train/Dev/Test" are original data without code-switched.}\label{tb:data_session}
\begin{tabular}{l|lll|llll|llll}

\toprule
 \multirow{2}{*}{Model} &\multicolumn{3}{c|}{AliExpress} &\multicolumn{4}{c|}{LAZADA} &\multicolumn{4}{c}{DARAZ}  \\\cmidrule{2-12}
                          & Ar   & En  & Zh  & Id   & Ms   & Fil   & Th   & Ur & Bn & Ne & Si\\
\midrule
Train  &  $12.8$K & $16.0$K & $11.2$K & $20.1$K & $18.8$K & $20.7$K & $20.7$K & $6.9$K & $8.7$K & $26.1$K & $4.0$K \\
Dev  & $1$K  & $1$K & $1$K & $1$K & $1$K & $1$K & $1$K & $1$K & $1$K & $2.0$K & $0.5$K \\
Test  & $1$K  & $1$K & $1$K & $1$K & $1$K & $1$K & $1$K & $1$K & $1$K  & $2.0$K & $0.5$K \\
\bottomrule

\end{tabular}
\end{table*}

\begin{table*}[!t]
\centering
\small
\caption{The Code-switching rate of each query for LAZADA. ``Mixed" stands for code-switched queries.}
\label{tb:code_mix_rate}
\begin{tabular}{l|lll|lll}

\toprule
 \multirow{2}{*}{Languages} &\multicolumn{3}{c|}{Code-switching Rate (Offline)} &\multicolumn{3}{c}{Code-switching Rate (Online)}  \\\cmidrule{2-7}
    & Mixed & English & Native & Mixed & English & Native \\
\midrule
\rowcolor{Gray}
Indonesian (Id) & $76.92$\%  & $1.16$\% & $21.92$\%   & $85.23$\% &  $0.34$\%  & $14.43$\% \\
Malay (Ms) & $27.90$\%  & $71.60$\% & $0.50$\% &  $38.87$\% & $57.06$\%  & $2.38$\%  \\
\rowcolor{Gray}
Filipino (Fil)  & $49.31$\%  & $50.60$\% & $0.08$\% &  $72.09$\% & $26.86$\%  & $1.04$\%  \\
Thai (Th) & $4.49$\%  & $1.84$\% & $93.67$\% &  $10.38$\% & $4.67$\%  & $84.95$\%  \\
\bottomrule

\end{tabular}
\end{table*}

\begin{table}[!t]
\centering
\small
\caption{Hyper-parameter settings.}
\begin{tabular}{l|l}

\toprule
Parameter                &   Value                  \\
\midrule
Word Embedding & $1280$ \\
Vocabulary Size & $200$K \\
Dropout & $0.1$ \\
Learning Rate & $1e-5$ \\
Margin & $0.1$ \\
Optimizer & Adam \\
Masking Probability & $0.15$ \\
$\lambda$ & 0.2 \\
Code-switching Rate & $10\%$ \\
\bottomrule

\end{tabular}
\label{table:par}
\end{table}

\subsection{Building Code-switched Data for SR} 

We aim to improve the SR model in the business scenario. Since the LAZADA corpus includes many code-switched sentences, we build the code-switched data using authentic business corpora and language features.
During the construction of the code-switched data, we replace each token among the $query$ and $label$ into the corresponding multi-lingual words based on some openly available multi-lingual lexicon-level dictionaries with some percentages.

\begin{table*}[htp]
\centering
\small
\caption{The comparison with Recall@30 between baseline systems on business corpora.}
\begin{tabular}{l|lll|llll|llll|l}

\toprule
 \multirow{2}{*}{Model} &\multicolumn{3}{c|}{AliExpress} &\multicolumn{4}{c|}{LAZADA} &\multicolumn{4}{c|}{DARAZ} &\multirow{2}{*}{Avg.}   \\\cmidrule{2-12}
                          & Ar   & En  & Zh  & Id   & Ms   & Fil   & Th   & Ur & Bn & Ne & Si &  \\
\midrule
 \textsc{mBERT} & 79.6 & 78.0 & 89.4 & 55.3 & 53.6 & 70.4 &  71.1  & 83.5 & 82.3 & 56.7&  75.8 & \cellcolor{Gray}72.3 \\
 \textsc{Unicoder} & 64.3 & 69.2 & 79.9 & 46.0 & 48.8 & 64.4 & 62.1 & 74.6 & 75.3 &  48.7 & 65.8 & \cellcolor{Gray}63.6 \\
 \textsc{XLMR}$_{Large}$ & 81.1 & 81.0 & 90.1 & 68.3 & 59.9 & 71.2 & 82.1 & 85.6 & 84.5 & 65.2 & 70.2 & \cellcolor{Gray}76.3\\
 \textsc{SimCSE-BERT}$_{Large}$  & 72.2 & 79.0  & 52.0 & 49.3 & 56.5 & 72.2 & 78.5 & 82.8 & 83.1  & 57.6  & 76.0 & \cellcolor{Gray}69.0 \\
 \textsc{InfoXLM}$_{Large}$ & 79.7 & 82.5 & 89.6 & 69.0 & 58.1 & 75.4 & 80.4 &  82.7 & 80.7   &  60.2  &  76.8 & \cellcolor{Gray}75.9 \\
 \textsc{VECO} & 85.9 & \textbf{83.0} & 91.4 & 68.7 & 58.7 & 75.3 & 82.3  & 87.3 & 87.6 & 61.4 & 81.4 & \cellcolor{Gray}{78.1} \\
 \textsc{CMLM} & 84.3 & 80.4 & 91.2 & 65.3 & 57.8 & 74.3 & 76.7 & 85.4 &  87.5 &  61.6 &  81.4 & \cellcolor{Gray}76.9\\
\textsc{LaBSE} & 85.8 & 81.1 & 91.4 &  65.8 & 59.8 & 75.7  & 76.7  &  85.7  &  85.6  & 62.9  & 81.6 & \cellcolor{Gray}77.5 \\
 \midrule
\rowcolor{Gray}
\textsc{Ours} & \textbf{89.0} & 82.6 & \textbf{93.9}  & \textbf{73.7} & \textbf{62.8} & \textbf{78.7}  &  \textbf{83.5}  &  \textbf{88.5}  &  \textbf{90.9}  &  \textbf{67.2} & \textbf{84.0} & \textbf{81.3} \\
\bottomrule

\end{tabular}
\label{tb:XLSR_business_data}

\end{table*}

The code-switched cross-lingual corpus consists of two parts such as $query$ and $label$. The original $query$ and $label$ are formulated as follows:
    \begin{align}
    \mathbf{q}&= \{q_1,\dots,q_i,\dots,q_I\}, \\
    \mathbf{l}&= \{l_1,\dots,l_j,\dots,l_J\},
    \end{align}
\noindent where the $I$ and $J$ represent the length of $query$ and $label$, respectively. We replace the tokens among the $query$ and $label$ with the frequently used languages in Alibaba over-sea's cross-border e-commerce platform. The newly constructed data should be as follow:
    \begin{align}
    \mathbf{q}'&= \{q_1^{\prime},\dots,q_i^{\prime},\dots,q_I^{\prime}\}, \\
    \mathbf{l}'&= \{l_1^{\prime},\dots,l_j^{\prime},\dots,l_J^{\prime}\},
    \end{align}
\noindent where $q_i^{\prime}$ and $l_j^{\prime}$ denote the tokens after the replacement (for all integers $i \in [1,I]$ and $j \in [1,J]$). As the final step, we combine the newly generated $\mathbf{q}'$ and $\mathbf{l}'$ to build the linguistically motivated code-switched monolingual corpus (i.e., $\langle \mathbf{q}',\mathbf{l}' \rangle$). Then we continually pre-train our model with a similar idea of ALM.

\section{Experiments}

\subsection{Setup}

\paragraph{Data preparation}
The languages selected from the business dataset are Arabic (Ar), English (En), Chinese (Zh), Indonesian (Id), Malay (Ms), Filipino (Fil), Thai (Th), Urdu (Ur), Bengali (Bn), Nepali (Ne), and Sinhala (Si). 
Specifically, Ar, En, and Zh are originated from AliExpress corpora, while Id, Ms, Fil, and Th are from the LAZADA corpora, and Ur, Bn, Ne, and Si are from DARAZ corpora, respectively. 
The characteristics of our business corpora are shown in Table \ref{tb:data_session}. 
Among them, the LAZADA corpus belongs to the code-switched dataset, and we provide the code-switching rates both on offline and online data separately (See Table \ref{tb:code_mix_rate}).
We also make some explorations on the SR task using the Quora Duplicate Questions Dataset\footnote{\url{https://quoradata.quora.com/First-Quora-DatasetRelease-Question-Pairs}} with Faiss \citep{faiss} toolkit\footnote{\url{https://github.com/facebookresearch/faiss}}.
Then we evaluate the model performance by exploiting the mean reciprocal rank (MRR) to validate the effectiveness of different approaches. 
Additionally, we conduct our experiments on the semantic textual similarity (STS) task using the SentEval toolkit \citep{infersent_emnlp17} for evaluation. 

For model robustness, we conduct an experiment on the openly available dataset AskUbuntu\footnote{\url{https://github.com/taolei87/askubuntu}} \citep{AskUbuntu} in English. 
We also make an investigation on the Tatoeba corpus \citep{laser_taacl19} in $11$ language pairs by exploiting the BUCC2018 corpus \citep{bucc} in $4$ language pairs, which are originated from the well-known and representative benchmark XTREME\footnote{\url{https://github.com/google-research/xtreme}} \citep{XTREME}.
In the STS task,  we leverage Spearman's rank correlation coefficient to measure the quality of correlation between human labels and calculated similarity \citep{SimCSE21}. 
We exploit the bilingual dictionary ConceptNet5.7.0 \citep{DBLP:conf/aaai/SpeerCH17} and MUSE \citep{DBLP:conf/iclr/LampleCRDJ18} during the generations of the code-switched data.
However, for the English corpus, we keep the original English sentences and do not leverage the code-switching.
We conduct all the experiments on Zh without Chinese word segmentation (for AliExpress) and without converting them into simplified scripts for BUCC corpora. The hyper-parameters as shown in Table \ref{table:par}.

\paragraph{Baselines}
To further verify the effectiveness of our method, we compare the proposed approach with the following highly related methods:

\begin{itemize}
    \item \textsc{mBERT} \citep{bert_naacl19} is transformer based multi-lingual bidirectional encoder representation and is pre-trained by leveraging the MMLM on the monolingual corpus.
    \item \textsc{Unicoder} \citep{unicoder_emnlp19} by taking advantage of multi-task learning framework to learn the cross-lingual semantic representations via monolingual and parallel corpora to gain better results on downstream tasks.
    \item  \textsc{XLM-R} \citep{xlm-r_acl2020} is more efficient than XLM and uses huge amount of mono-lingual datasets that originated from Common Crawl \citep{common_crawl_dataset} which includes 100 languages to train MMLM.
    \item  \textsc{SimCSE} \citep{SimCSE21} propose a self-predictive contrastive learning that takes an input sentence and predicts itself as the objective.
    \item \textsc{InfoXLM} \citep{info-xlm_naacl21} is an efficient method to learn the cross-lingual model training by adding a constraints.
    \item \textsc{VECO} \citep{veco_acl2021} obtains better results on both generation and understanding tasks by introducing the variable enc-dec framework.
    \item \textsc{CMLM} \citep{universal_sen_rep2020} is a totally unsupervised learning method, conditional MLM, can effectively learn the sentence representation on huge amount of unlabeled data via integrating the sentence representation learning into MLM training.
    \item \textsc{LaBSE} \citep{labse_acl22} adapts the mBERT to generate the language-agnostic sentence  embedding for 109 languages and is pre-trained by combining the MLM and TLM with translation ranking task leveraging bi-directional dual encoders. 
\end{itemize}

\subsection{Main Results}

\subsubsection*{SR Results on Business Data}
\begin{figure}[!t]
\centering
\includegraphics[width=7.7cm,height=4.5cm]{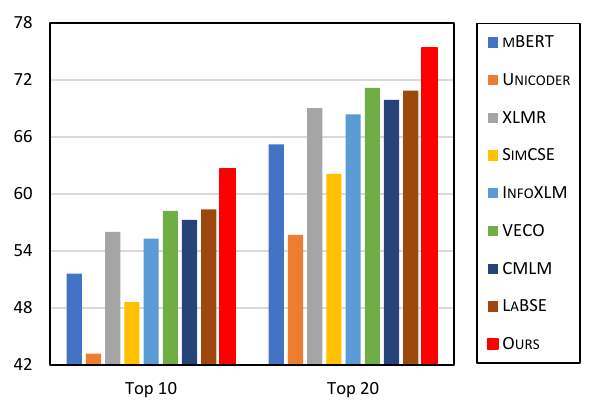}
\caption{\label{fig:top10_20} The comparison with average Recall@\{10,20\} on the business dataset.}
\end{figure}

Table \ref{tb:XLSR_business_data} shows the retrieving results of the proposed method on Ali-Express, DARAZ, and LAZADA corpora by evaluating the Recall@30 score, respectively.
The conduction of our experiment is composed of two steps, firstly we continually pre-train the models by combining the queries and labels among the training set. Then we conduct the fine-tuning on different languages by exploiting their own train set and dev set.
Unlike other baselines, in our continual pre-training step, we utilize code-switched queries and labels instead of combining the original data to train our model.
Among the baselines, \textsc{VECO} achieves better results on almost every language from the business corpus. 
The code-switching method has the most positive effects both on Id, Fil (The corpora of these two languages are highly code-switched. See Table \ref{tb:code_mix_rate}) and Bn, Si (DARAZ), but brings fewer benefits for Zh (Ali-Express). As we do not leverage any code-switched data for En, we obtain less improvement than VECO. 
However, our approach consistently outperforms all the baselines on each language except En.
As depicted in Figure \ref{fig:top10_20}, we also evaluate the models with Recall@\{10,20\} (For the details, see Table \ref{tb:XLSR_business_data_top10} \& \ref{tb:XLSR_business_data_top20} in Appendix).  

\subsubsection*{Results of Semantic Textual Similarity (STS)}

\begin{figure}[!t]
\centering
\includegraphics[width=7.5cm,height=4.5cm]{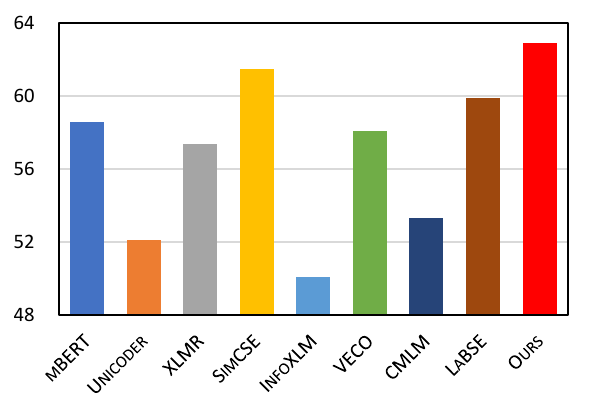}
\caption{\label{fig:sts} 
The comparison of sentence embedding performance with \textbf{average} Spearman's rank on \textbf{STS tasks}.}
\end{figure}

As depicted in Figure \ref{fig:sts}, we further verify the model performance of our proposed approach on the highly similar task STS that is close to SR. 
In this experiment, all of the test sets only include English sentences. Thus we continually pre-train each baseline on the \textbf{BUCC2018} corpora by combing all the English monolingual datasets. 
For a fair comparison, we exploit the BUCC data to continually pre-train our model. 
We evaluate the baselines and our model on the test sets only using the continually pre-trained model instead of the fine-tuned model.
The presented model also obtains consistent improvements on all test sets. 
We provide more details in Table \ref{tb:sts_task_result} (see Appendix).

\subsubsection*{Verification of Robustness}
As shown in Table \ref{tb:XLSR_open_data}, we also further explore the performance of our model on the openly available corpus \textbf{AskUbuntu} \citep{AskUbuntu} and \textbf{Tatoeba} benchmark \citep{laser_taacl19}.
In this experiment, we conduct continual pre-training and fine-tuning by leveraging only the AskUbuntu corpus without using other datasets. 
We evaluate all the baselines and our model using different evaluation metrics P@1, P@5, and R@30 with \textsc{Top-1}, \textsc{Top-5}, and \textsc{Top-30} queries, respectively.  
Moreover, we further verify the effectiveness of our approach on another openly available benchmark Tatoeba and obtain remarkably better results than baselines. For more details, see Table \ref{tb:tatoeba_details} in Appendix.

\subsection{Monolingual Semantic Retrieval}
As shown in Table \ref{tb:mono_lin_SR}, we also tend to verify the semantic retrieving skill of our approach on another openly available \textbf{Quora Duplicate Questions} dataset, which only includes English monolingual data.
Since this corpus only provides the test set, we merge the English part of the train set from BUCC18 as our monolingual data. Then we continually pre-train each model to evaluate them on the Quora Duplicate Questions dataset without using the fine-tuned model.
In this dataset, \textsc{InfoXLM}$_{Large}$ obtains better results than other baselines, but our method outperforms all the baselines, which indicates that our approach has better retrieving skills compared to similar methods.

\begin{table}[!t]
\centering
\small
\caption{The comparison with various evaluation metric on \textbf{AskUbuntu} corpus. ``P@1" and ``P@5" denote the precision score on \textsc{\textbf{Top-1}} query and \textsc{\textbf{Top-5}} queries, respectively. ``R@30'' denotes the recall score on \textsc{\textbf{Top-30}} queries.}
\begin{tabular}{l|cc|c}

\toprule
 \multirow{2}{*}{Model} &\multicolumn{3}{c}{AskUbuntu}  \\\cmidrule{2-4}
                        & P@1 & P@5 & R@30 \\
\midrule

 \textsc{mBERT}  & 49.0 &  41.2 & 54.6 \\
 \textsc{Unicoder}  & 52.2 & 38.1 & 45.5 \\
 \textsc{XLMR}$_{Large}$  &  55.4 & 43.4 & 60.1 \\
  \textsc{SimCSE-BERT}$_{Large}$ & 53.2 & 42.6 & 56.1 \\
 \textsc{InfoXLM}$_{Large}$  & 52.7 & 43.2 & 55.8 \\
 \textsc{VECO}  & 53.2 & 41.8 & 59.8 \\
 \textsc{CMLM} & 53.2 &  41.0 & 59.6 \\
\textsc{LaBSE} &  54.8 & 42.8 & 59.3 \\
 \midrule
\rowcolor{Gray}
\textsc{Ours} & \textbf{57.5} & \textbf{43.8} & \textbf{61.1} \\
\bottomrule

\end{tabular}
\label{tb:XLSR_open_data}
\end{table}

\subsection{Ablation Study}

\subsubsection*{The Effect of Similarity Loss $\mathcal{L}^{(\mathbf{q},\mathbf{l})}_{sim}$}

As illustrated in Figure \ref{fig:lambda_and_cdm}(a), it is an essential part of the cross-lingual PTM with similarity. We observe that the similarity brings a positive effect on the performance of our model.
Our approach achieves better improvements 
with learning the similarity loss during the pre-training stage than without similarity compared with other baselines.

\subsubsection*{The Effect of $\lambda$}

$\lambda$ controls the weight of the XMLM, which appears in Equation (\ref{eq_joint_loss}). As depicted in Figure \ref{fig:lambda_and_cdm}(b), when $\lambda = 0.2$, our model achieves the best retrieving performance compared with other values. 
We provide the details of the effectiveness of different values for $\lambda $ in Table \ref{tb:dif_lmda} (see Appendix).

\subsubsection*{The Effect of Code-switching}

\begin{figure*}[!t]
    \centering
    \begin{subfigure}[b]{0.48\textwidth}
        \centering
        \includegraphics[scale=0.29]{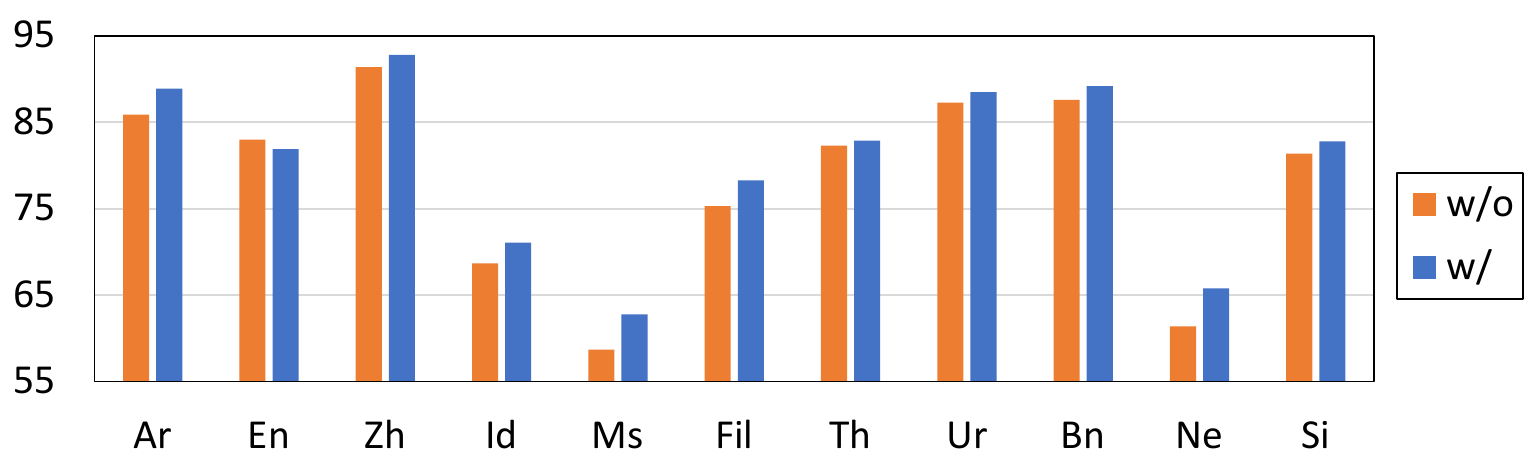}
        \caption{\textsc{with and without Similarity Loss}}
        \label{fig:sim_loss}
    \end{subfigure}
    \begin{subfigure}[b]{0.25\textwidth}
        \centering
        \includegraphics[scale=0.29]{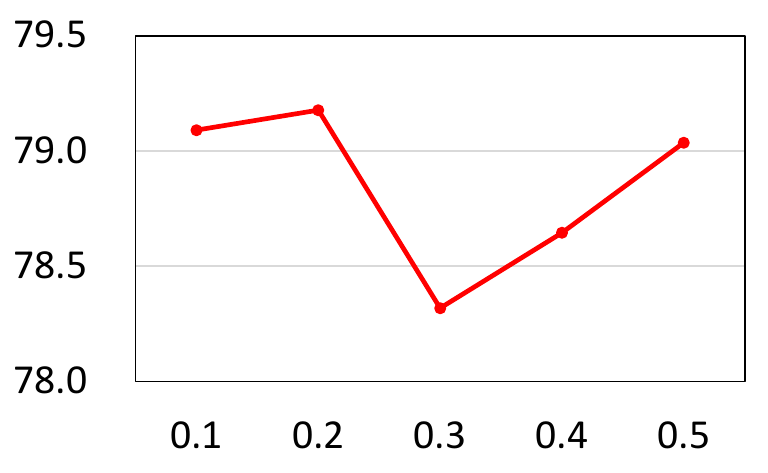}
        \caption{\textsc{Value of $\lambda$}}
        \label{fig:lambda}
    \end{subfigure}
    \begin{subfigure}[b]{0.25\textwidth}
        \centering
        \includegraphics[scale=0.29]{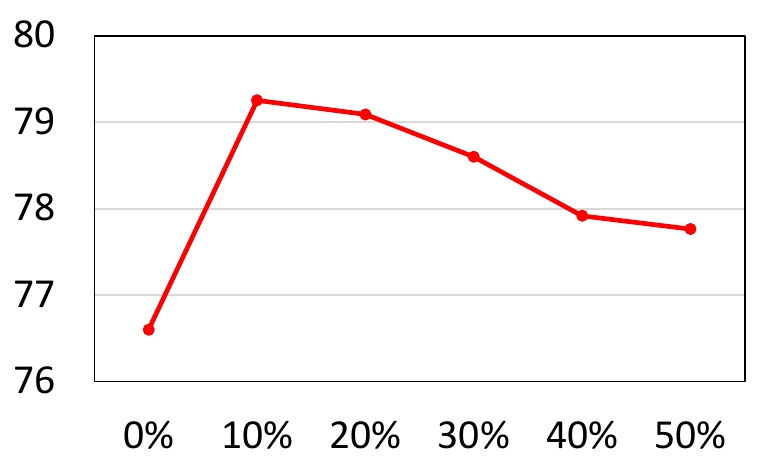}
        \caption{\textsc{Value of $Cmd_r$}}
        \label{fig:cdm}
    \end{subfigure}
    
    \caption{The effect of similarity loss ($\mathcal{L}^{(\mathbf{q},\mathbf{l})}_{sim}$) and different values of the hyper-parameters $\lambda$  and \textsc{Code-mixing Rate} ($Cmd_r$) in our model on business corpora with Recall@30. (a) ``w/" and ``w/o" denote the accuracy score with or without similarity loss. (b) and (c) also represent the \textbf{average Recall@30} score with different values of $\lambda$ (default value is $0.2$) and $Cmd_r$ (default value is $10\%$), respectively.
    }
    
    \label{fig:lambda_and_cdm}
\end{figure*}

As illustrated in Figure \ref{fig:lambda_and_cdm}(c), we also investigate the effectiveness of code-switching for our method.
First, the performance becomes lower if we keep the data without code-switching ($Cmd_r = 0\%$), which demonstrates the effectiveness of code-switching. Since all the languages in the LAZADA corpus originally included code-switched scripts, it may obtain lower performance when we train the model without code-switched data.
Second, when $Cmd_r = 10\%$, our model reaches the best average performance (For more details, see Table \ref{tb:dif_cmd_rate} in Appendix).

\begin{table}[!t]
\centering
\small
\caption{The comparison with MRR@10 and Recall@30 evaluation metrics on the \textbf{Quora Duplicate Questions} dataset.}
\begin{tabular}{l|r|r}

\toprule
Model &  MRR@10 & Recall@30\\
\midrule
 \textsc{mBERT} & 0.436 & 0.506 \\
 \textsc{Unicoder} & $0.227$ & 0.253 \\
 \textsc{XLMR}$_{Large}$ &  $0.498$ & 0.547 \\
 \textsc{SimCSE-BERT}$_{Large}$ & 0.453 & 0.538 \\
 \textsc{InfoXLM}$_{Large}$ & 0.575 & 0.620\\
 \textsc{VECO} & $0.517$ & 0.579\\
 \textsc{CMLM} & 0.551 & 0.605 \\
\textsc{LaBSE} & 0.493 & 0.560 \\
 \midrule
 \rowcolor{Gray}
\textsc{Ours} & \textbf{0.584} & \textbf{0.644 }\\
\bottomrule

\end{tabular}
\label{tb:mono_lin_SR}

\end{table}

\section{Related Work}

\paragraph{Semantic Retrieval}

Semantic retrieval is an essential task in NLP, which requires the model to calculate the sentence embeddings, and then similar sentences can be retrieved by the embeddings. 
Early SR methods are constructed based on traditional word2vec representations \citep{skipthought_nips15,sen_rep_naacl16}. Subsequently, various studies have proposed using siamese networks to perform semantic retrieval~\citep{neculoiu2016learning,kashyap2016robust,wei_2018_ialp}.
With widely using the pre-trained language models,
Reimers and Gurevych \citep{sbert2019} propose Sentence-BERT, which learns sentence embeddings by fine-tuning a siamese BERT network on NLI datasets. 
To conduct multi-lingual SR, Reimers and Gurevych \citep{knowledgedis_emnlp20} extend Sentence-BERT to its multi-lingual version by knowledge distillation. 
To better leverage unlabeled data for SR, \citet{SimCSE21} propose SimCSE, which uses contrastive learning to train sentence embeddings. 
To improve performances of multi-lingual SR, \citet{info-xlm_naacl21} propose InfoXLM, which utilizes MLM, TLM, and contrastive learning objectives. 

\paragraph{Code-switching}

Code-switching is a pre-training technique to improve cross-lingual pre-trained models. That is used in PTMs for machine translation \citep{csp_emnlp20,ptmnmt_emnlp20,alm2020aaai}.
For example, 
\citet{csp_emnlp20} utilize code-switching on monolingual data by replacing some continuous words into the target language and letting the model predict the replaced words. 
\citet{ptmnmt_emnlp20} use code-switching on the source side of the multi-lingual parallel corpora to pre-train an encoder-decoder model for multi-lingual machine translation. 
\citet{Feng2022TowardTL} mitigate the limitation of the code-switching method for grammatical incoherence and negative effects on token-sensitive tasks.
\citet{alm2020aaai} propose ALM for cross-lingual pre-training, which requires the model to predict the masked words in the code-switched sentences. \citet{Krishnan2021MultilingualCF} augment monolingual source data by leveraging the multilingual code-switching via random translation to improve the generalizability of large multi-lingual language models.
Besides, code-switching has been utilized in other NLP tasks, including named entity recognition \citep{cdmx_ner18}, question answering \citep{cdmx_qa18,cdmx_qa_mt18}, 
universal dependency parsing \citep{cdmx_parsing_naacl18}, morphological tagging \citep{cdmx_morph_tag21},
language modeling \citep{cdmx_lm18}, automatic speech recognition \citep{Kumar2018PartofSpeechAO}, natural language inference \citep{cdm_nli_2020} and sentiment analysis \citep{semevl_tsk9,cdmx_sa20,semevl20_tsk9,Zhang2021CrosslingualAS}.
To the best of our knowledge, we are the first to utilize code-switching for semantic retrieval.

\section{Conclusion and Future work}

We introduce a straightforward pre-training approach to sentence-level semantic retrieval with code-switched cross-lingual data for the FAQ system in the task-oriented QA dialogue e-commerce scenario. 
Intuitively, code-switching is an emerging trend of communication in both bilingual and multi-lingual regions.
Our experimental result shows that the proposed approach remarkably outperforms the previous highly similar baseline systems on the tasks of semantic retrieval and semantic textual similarity with three business corpora and four open corpora using many evaluation metrics. 
In future work, we will expand our method to other natural language understanding tasks. Besides, we will also leverage the different embedding distance calculation metrics instead of only using cosine similarity.

\section*{Acknowledgments}

We are deeply thankful to each of our esteemed co-authors for their precious contributions and perceptive comments, which greatly improved the standard of this paper before its submission. Their knowledge and cooperative attitude were vital in forming our research. Moreover, our appreciation goes to our peers who carefully examined our work and identified minor mistakes, guaranteeing the precision and excellence of our conclusions. Their meticulous attention to detail substantially bolstered the credibility of our investigation.

\bibliography{acl_latex}
\bibliographystyle{acl_natbib}

\clearpage
\appendix

\begin{figure*}[htbp]
    \centering
    \begin{subfigure}[t]{0.44\textwidth}
        \centering
        \includegraphics[trim={2.5cm -1.5cm 2cm 0cm},clip,scale=0.2]{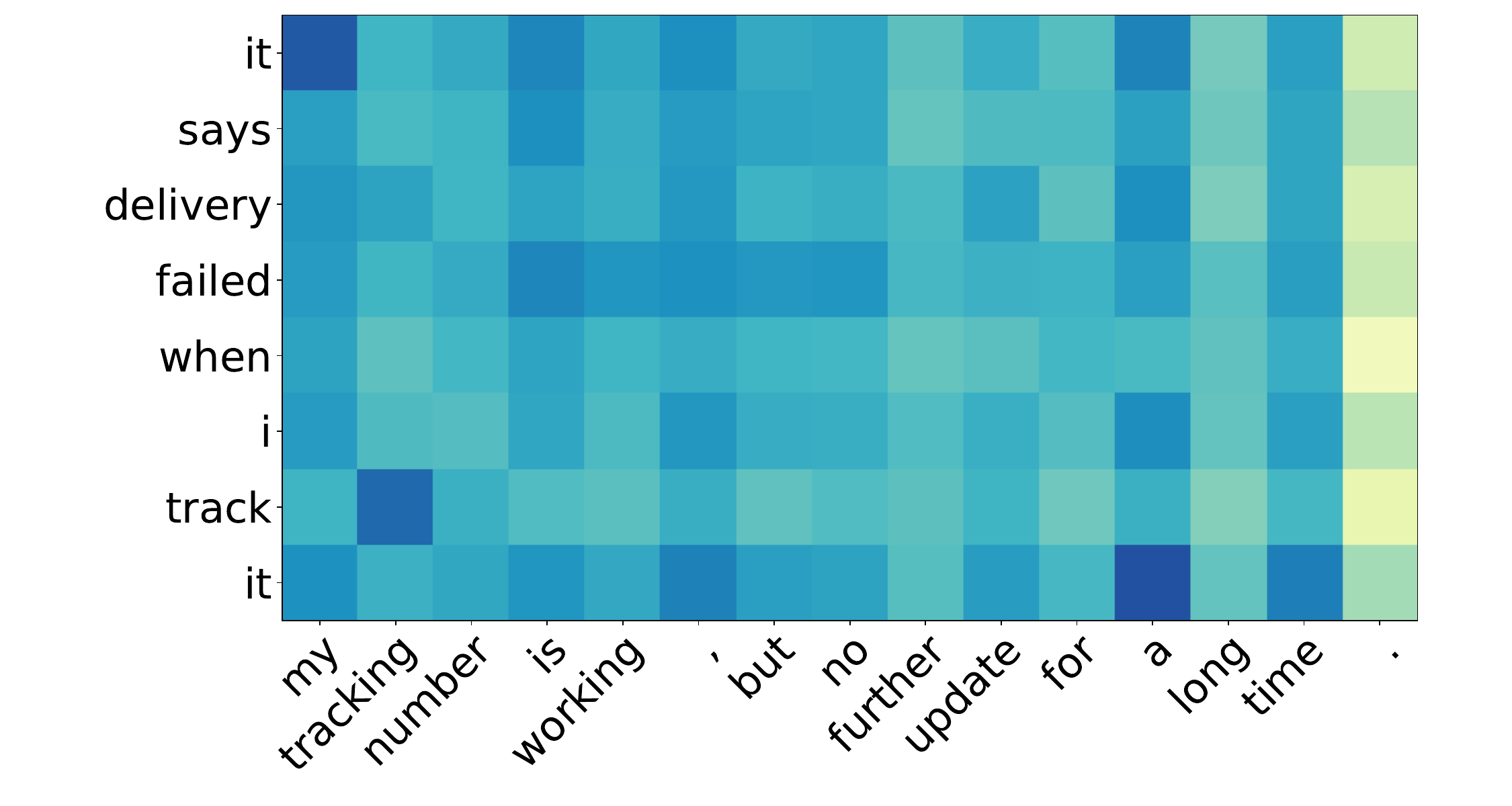}
        \caption{\textsc{CMLM}}
    \end{subfigure}
    \begin{subfigure}[t]{0.27\textwidth}
        \centering
        \includegraphics[trim={1cm 0.5cm 0cm 0cm},clip,scale=0.22]{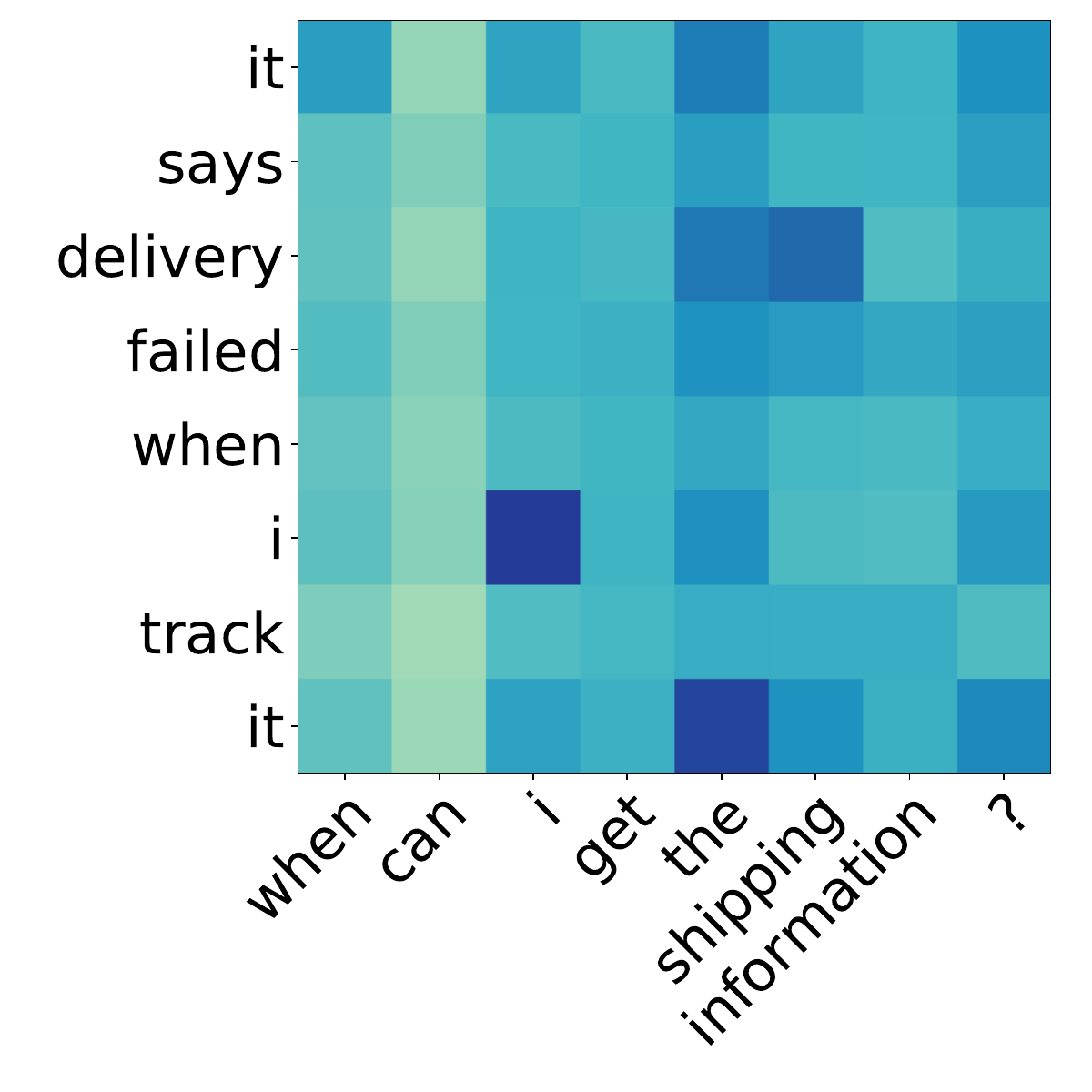}
        \caption{\textsc{XLMR}}
    \end{subfigure}
    \begin{subfigure}[t]{0.27\textwidth}
        \centering
        \includegraphics[trim={1cm -0.5cm 0cm 0cm},clip,scale=0.22]{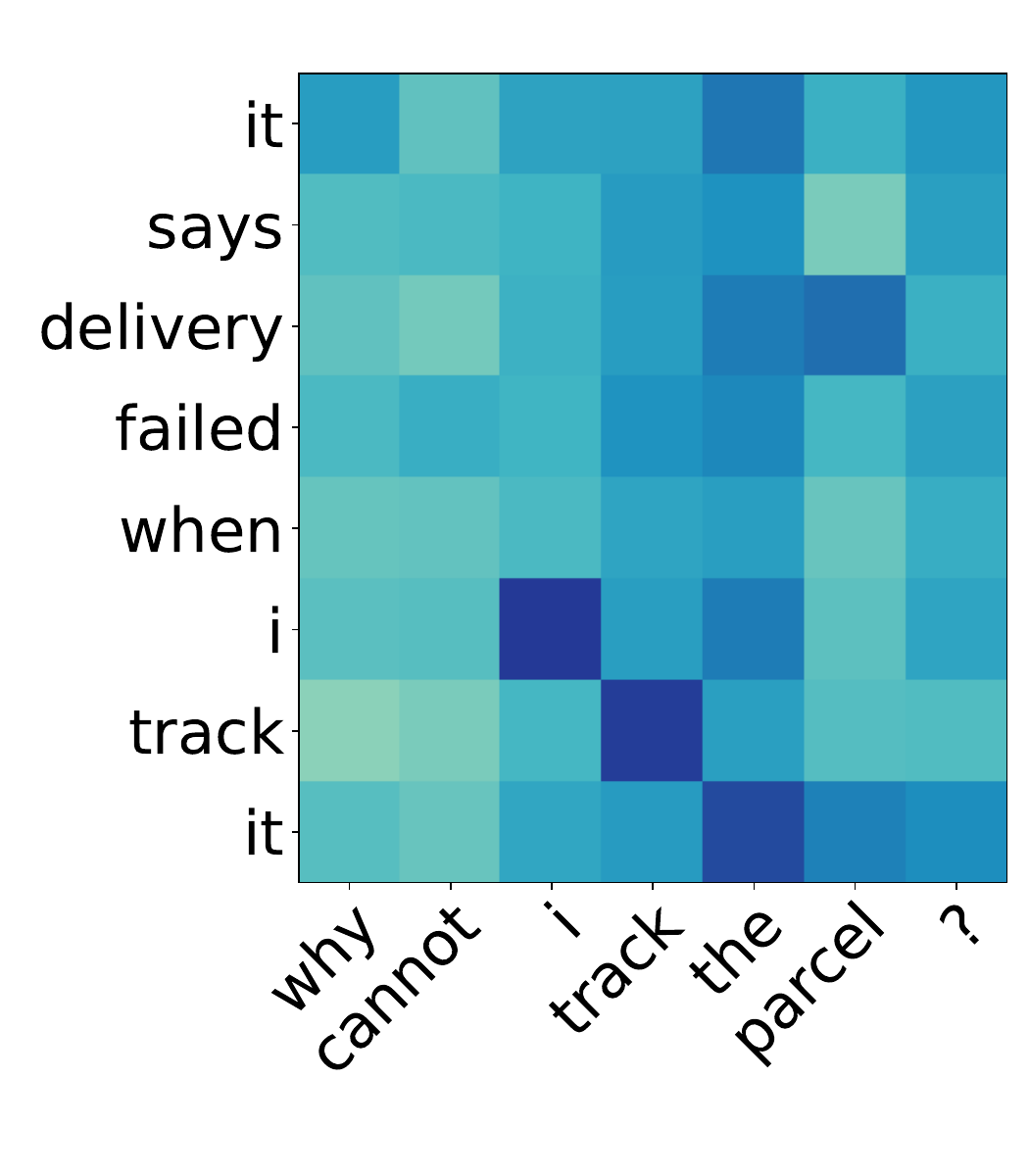}
        \caption{\textsc{LaBSE}}
    \end{subfigure}
    \\
    \begin{subfigure}[t]{0.44\textwidth}
        \centering
        \includegraphics[trim={2.5cm 0cm 2cm 0cm},clip,scale=0.2]{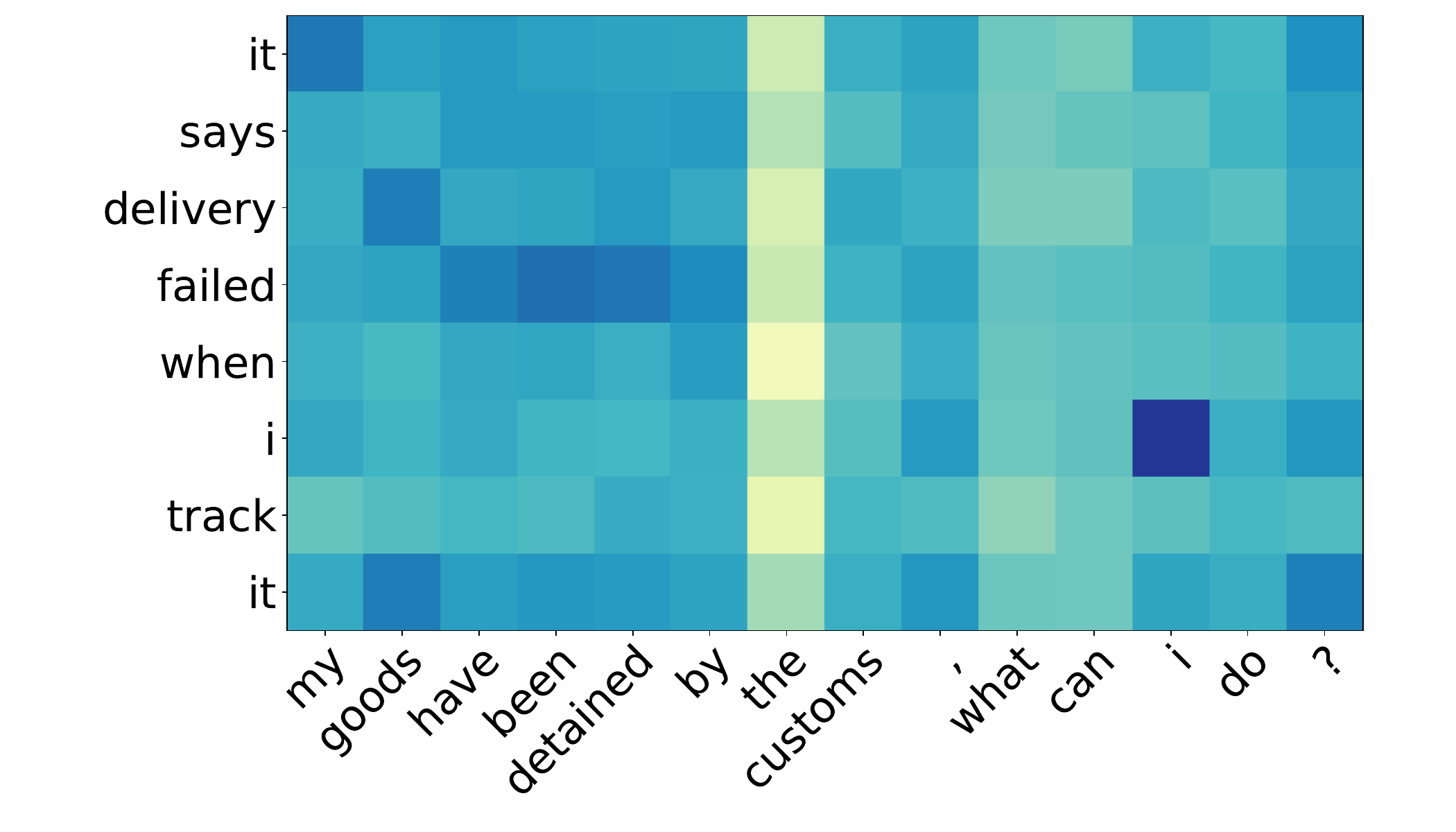}
        \caption{\textsc{InfoXLM}}
    \end{subfigure}
    \begin{subfigure}[t]{0.27\textwidth}
        \centering
        \includegraphics[trim={0cm 0cm 0cm 0cm},clip,scale=0.21]{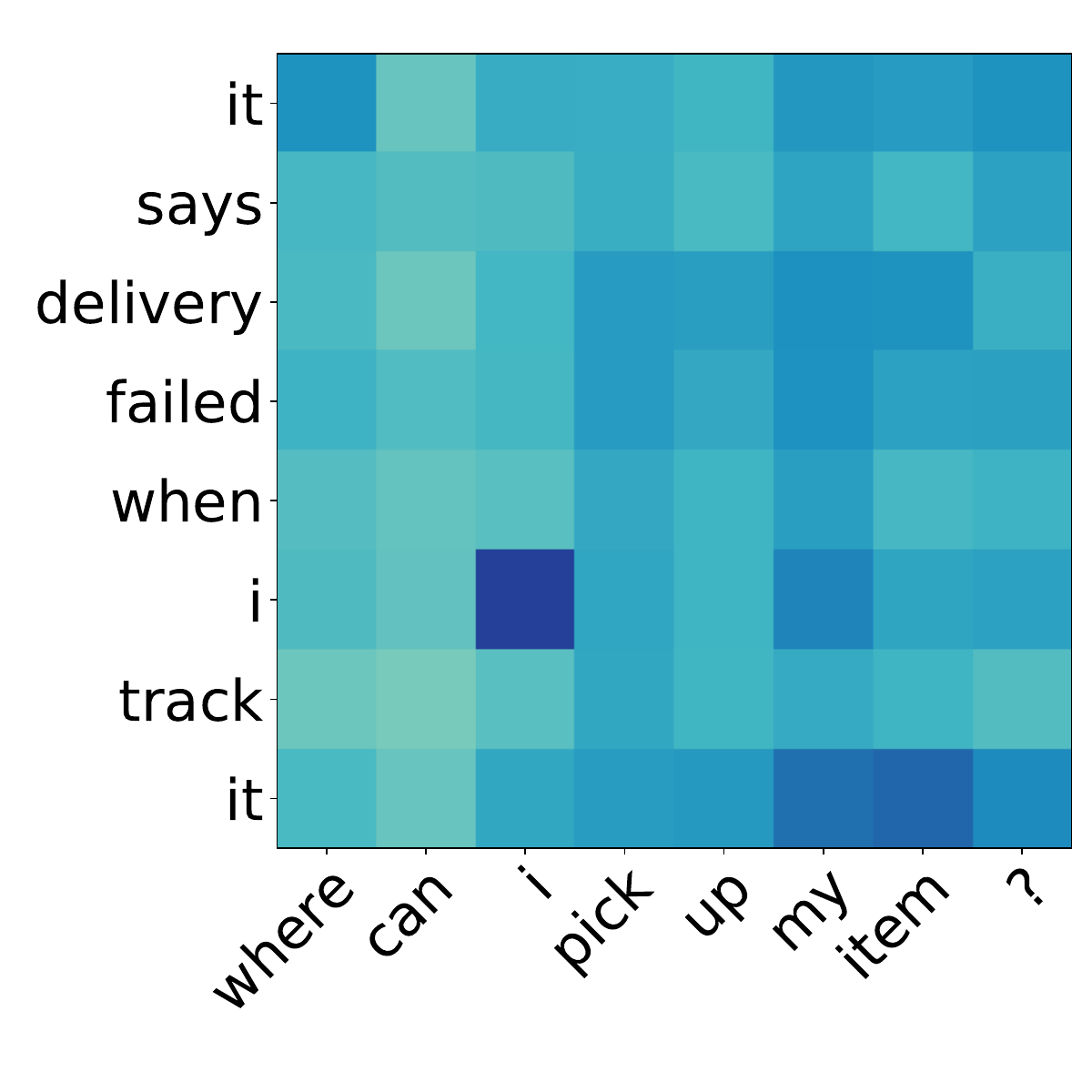}
        \caption{\textsc{VECO}}
    \end{subfigure}
    \begin{subfigure}[t]{0.27\textwidth}
        \centering
        \includegraphics[trim={0cm 2cm 0cm 0cm},clip,scale=0.24]{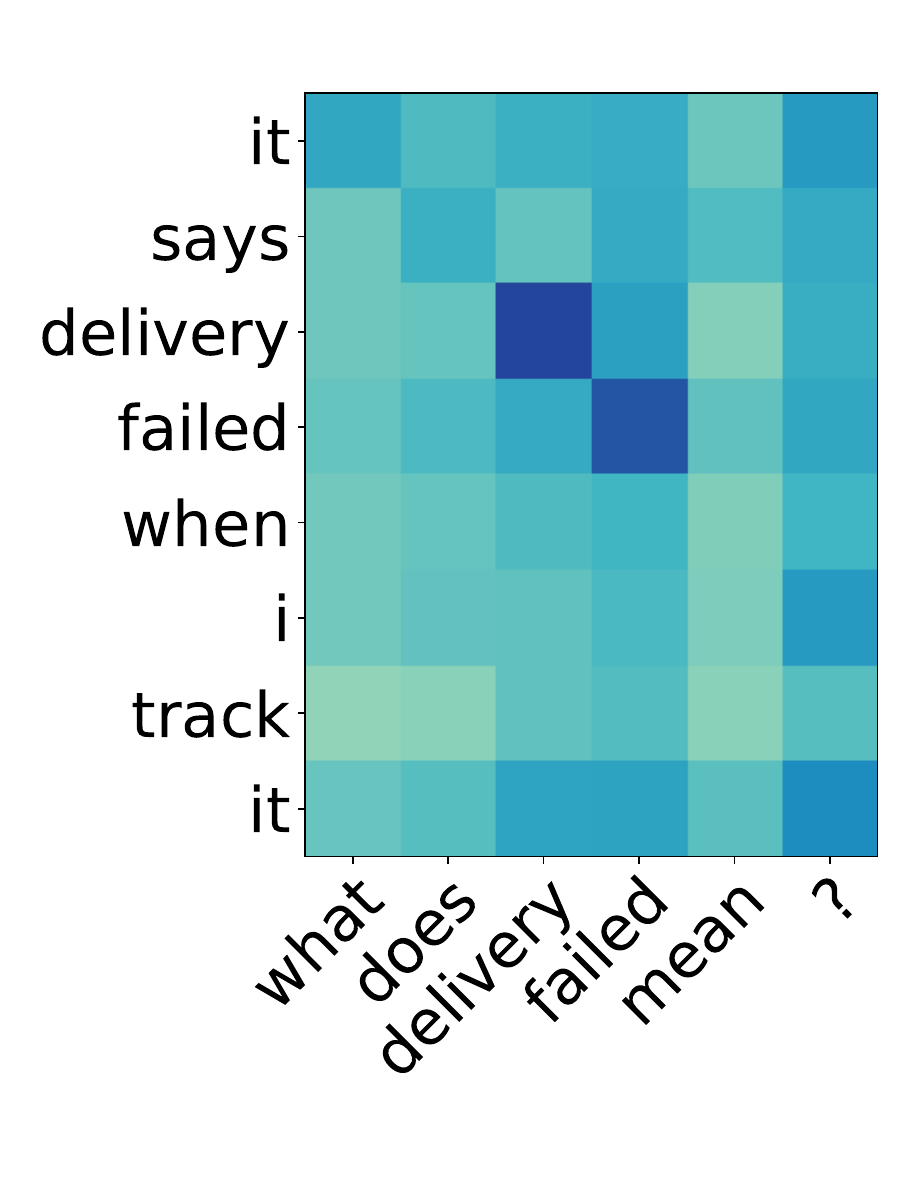}
        \caption{\textsc{Ours}}
    \end{subfigure}
    \caption{The comparison of the retrieved labels among the baselines and our proposed method, where the y-axis and the x-axis denote the user query and the retrieved label, respectively.}
    \label{fig:casestudy}
\end{figure*}

\begin{table*}[!t]
\centering
\small
\caption{The comparison with Recall@10 between cross-lingual sentence retrieval baseline systems on AliExpress, LAZADA, and DARAZ corpora.
}
\begin{tabular}{l|lll|llll|llll|l}

\toprule
 \multirow{2}{*}{Model} &\multicolumn{3}{c|}{AliExpress} &\multicolumn{4}{c|}{LAZADA} &\multicolumn{4}{c|}{DARAZ} &\multirow{2}{*}{Avg.}   \\\cmidrule{2-12}
                          & Ar   & En  & Zh  & Id   & Ms   & Fil   & Th   & Ur & Bn & Ne & Si &  \\
\midrule
 \textsc{mBERT} & 56.3 & 55.2 & 78.0 & 32.1 & 29.8 & 46.9 & 48.1 & 67.6  & 67.4  & 35.0 & 51.6 & \cellcolor{Gray}51.6 \\
 \textsc{Unicoder} & 43.1 & 49.8 & 66.4 & 25.2 & 24.7 & 41.7 & 41.9 & 59.3 & 56.5 & 27.9 & 38.8 & \cellcolor{Gray}43.2 \\
 \textsc{XLMR}$_{Large}$ & 65.1 & 61.0 & 79.1 & 45.7 & 36.0 & 47.1 & 59.4 & 68.2 & 66.5 & 43.5 & 44.6 & \cellcolor{Gray}56.0 \\
 \textsc{SimCSE-BERT}$_{Large}$  & 48.4 & 58.8 & 40.3 & 27.4 & 31.5 & 48.5 & 58.1 & 66.0 &  68.8 & 36.9 & 50.2 & \cellcolor{Gray}48.6 \\
 \textsc{InfoXLM}$_{Large}$ & 60.2 & 61.7 & 76.9 & 44.7 & 33.8 & 51.6 & 61.3 &  66.5 &  64.6 & 37.3 & 49.4 & \cellcolor{Gray}55.3 \\
 \textsc{VECO} & 67.4 & 61.7 & 81.7 & 47.6 & 35.3 & 52.8 & 60.0 & 70.7 & 70.6 & 40.3 & 52.4 & \cellcolor{Gray}58.2\\
 \textsc{CMLM} & 67.6 & 62.9 & 80.1 & 41.2 & 35.8 & 49.0 & 55.9 & 71.3 & 72.0 & 38.4 & 56.0 & \cellcolor{Gray}57.3\\
\textsc{LaBSE} & 71.1 & 61.3 & 81.0 & 41.5 & 35.8 & 52.1 & 55.5 & {71.8} & 74.1 & 40.7 & 57.4 & \cellcolor{Gray}{58.4} \\
 \midrule
 \rowcolor{Gray}
\textsc{Ours}& \textbf{73.1} & \textbf{64.5} & \textbf{85.1} & \textbf{50.1} & \textbf{38.4} & \textbf{56.6} & \textbf{62.2} &  \textbf{75.1} & \textbf{75.8} & \textbf{47.8} & \textbf{59.6} & \textbf{62.6} \\
\bottomrule

\end{tabular}

\label{tb:XLSR_business_data_top10}

\end{table*}

\begin{table*}
\centering
\small
\caption{The comparison with Recall@20 between cross-lingual sentence retrieval baseline systems on AliExpress, LAZADA, and DARAZ corpora.
}
\begin{tabular}{l|lll|llll|llll|l}

\toprule
 \multirow{2}{*}{Model} &\multicolumn{3}{c|}{AliExpress} &\multicolumn{4}{c|}{LAZADA} &\multicolumn{4}{c|}{DARAZ} &\multirow{2}{*}{Avg.}   \\\cmidrule{2-12}
                          & Ar   & En  & Zh  & Id   & Ms   & Fil   & Th   & Ur & Bn & Ne & Si &  \\
\midrule
 \textsc{mBERT} & 71.8 & 71.8 & 85.5 & 46.1 & 42.2 & 62.2 & 63.3 & 78.2 & 78.6 & 48.2 & 69.2 & \cellcolor{Gray}65.2 \\
 \textsc{Unicoder} & 55.6 & 63.0 & 74.2 & 36.9 & 37.3 & 56.4 & 54.6  & 69.0 & 68.6 & 40.7 & 56.6 & \cellcolor{Gray}55.7\\
 \textsc{XLMR}$_{Large}$ & 74.1 & 73.8 & 87.2 & 60.9 & 49.4 & 62.3 & {74.2} & 80.3 & 79.5 & 57.6 & 60.2 & \cellcolor{Gray}69.0 \\
 \textsc{SimCSE-BERT}$_{Large}$  & 63.3 & 71.9 & 47.5 & 40.6 & 47.2 & 64.4 & 73.7 & 78.4 & 79.2 & 50.6 & 66.6 & \cellcolor{Gray}62.1\\
 \textsc{InfoXLM}$_{Large}$ & 73.4 & 75.8 & 85.8 & 60.9 & 45.9 & 66.7 & 72.6 & 77.4 & 75.9 & 51.1 & 66.8 & \cellcolor{Gray}68.4\\
 \textsc{VECO} & 80.4 & {76.3} & 88.3 & {61.1} & 47.9 & {68.1} & 73.8  & {82.4} & 82.3 & 53.4 & 69.0 & \cellcolor{Gray}{71.2} \\
 \textsc{CMLM} & 80.3 & 74.9 & 88.4 & 56.4 & 47.1 & 64.9 & 69.0 & 80.9 & 82.8 & 52.5 & 72.0 & \cellcolor{Gray}69.9 \\
\textsc{LaBSE} & 82.2 & 73.6 & 88.3 & 58.4  & {49.5} & 67.3 & 68.8 & 81.4 & 81.1 & {54.9} & {74.2} & \cellcolor{Gray}70.9 \\
 \midrule
 \rowcolor{Gray}
\textsc{Ours}& \textbf{85.3} & \textbf{77.2} & \textbf{91.4} & \textbf{67.3} & \textbf{52.3} & \textbf{71.4} & \textbf{76.7} & \textbf{85.1} & \textbf{86.3} & \textbf{60.0} & \textbf{76.4} & \textbf{75.4} \\
\bottomrule

\end{tabular}

\label{tb:XLSR_business_data_top20}

\end{table*}

\begin{table*}[!t]
\centering
\small
\caption{The comparison of sentence embedding performance on \textbf{STS tasks}. ``STS12-STS16", ``STS-B" and ``SICK-R" denote SemEval2012-2016, STS benchmark and SICK relatedness dataset, respectively.}
\begin{tabular}{l|lllllll|l}

\toprule
Model     & STS$12$ & STS$13$ & STS$14$   & STS$15$   & STS$16$   & STS-B   & SICK-R & Avg. \\
\midrule
 \textsc{mBERT} & 42.46 & 62.14 & 52.35 & 65.36 & 66.20 & 60.51 & 60.87 & 58.56  \\
 \textsc{Unicoder} & $41.07$ & $56.92$ & $49.76$ & $60.86$ & $53.65$ & $47.97$ & $54.78$ &  $52.14$   \\
 \textsc{XLMR}$_{Large}$ & $39.79$ & $62.65$ & $52.09$ & $62.26$ & $64.39$ & $59.27$ & $61.07$ & $57.36$  \\
 \textsc{SimCSE-BERT}$_{Large}$ & {42.35} & {67.34} & \textbf{57.20} & {70.36 }& \textbf{69.41} & 59.86 & 63.77 & {61.47} \\
\textsc{InfoXLM}$_{Large}$  & 32.23  & 52.35 & 39.42 & 52.04 & 60.82 & 54.04 & 59.61 & 50.07 \\
 \textsc{VECO} & 41.76 & 60.75 & 52.21 & 64.79 & 67.26 & 58.93 & 61.17 & 58.12 \\
 \textsc{CMLM} & 30.14 & 61.77 & 47.45 & 61.25 & 62.73 & 53.23  &   56.62 &   53.31 \\
\textsc{LaBSE} & 47.43 & 64.13 & 55.72 & 69.66 & 64.21 & 57.60 & 60.68 & 59.92 \\
 \midrule
 \rowcolor{Gray}
\textsc{Ours} & \textbf{44.53} & \textbf{68.20} & {55.99} & \textbf{71.39} & {66.07} & \textbf{66.32} & \textbf{68.03}  &  \textbf{62.93} \\
\bottomrule

\end{tabular}
\label{tb:sts_task_result}

\end{table*}

\appendices
\section{Case Study}

To further demonstrate the better performance of the proposed approach, we make some visualizations of the retrieved results on the Business corpus between the previously introduced SOTA pre-trained models (i.g., we choose the five most efficient baselines) for cross-lingual scenarios.

As illustrated in Figure \ref{fig:casestudy}, the baselines fail to retrieve the key information ``delivery failed'' from the user query ``it says delivery failed when I track it'', while our proposed method can retrieve such vital information, which indicates that our model can improve the performance of sentence retrieval in the business domain.

\begin{table*}[!t]
\centering
\small
\caption{The comparison with Recall@30 on \textbf{Tatoeba} corpus for each language.}
\begin{tabular}{l|ccccccccccc}

\toprule
Model & af & de & es & fr & it & ja & kk & nl & pt & sw & te \\
\midrule
\textsc{mBERT} \citep{bert_naacl19} & 55.4 & 78.0  &  74.2 & 74.7 & 73.6 & 73.1 & 50.1 & 71.2 & 76.6 & 32.3 & 58.5   \\
\textsc{Unicoder} \citep{unicoder_emnlp19} & 15.8 & 15.4 & 20.2 & 40.4 & 19.4 & 32.6 & 16.5 & 21.7 & 20.7 & 25.4 & 18.4 \\
\textsc{XLMR} \citep{info-xlm_naacl21} & 67.8 & 86.6 & 86.2 & 87.7 & 83.2 & 84.9 & 46.4 & 84.2 & 86.8 & 36.7 & 70.9 \\
\textsc{SimCSE} \citep{SimCSE21} & 23.0 & 28.4 & 25.1 & 39.1 & 25.4 & 21.6 & 20.4 & 20.7 &  21.2 & 16.7 & 17.7 \\
\textsc{InfoXLM} \citep{info-xlm_naacl21} & 84.7 & 90.6 & 86.4 & 93.7 & 89.5 & 88.2 & 77.2 & 91.3 & 89.8 & 68.9 & 74.3 \\
\textsc{VECO} \citep{veco_acl2021} & 86.3 & {96.5} & {95.7} & 96.4 & {95.3} & {93.5} & 71.1 & 94.6 & 95.4 & 73.0 & 81.1 \\
\textsc{CMLM} \citep{universal_sen_rep2020} & 87.9 & 94.1 & 91.5 & 97.1 & 94.3 & 91.3 & 75.6 & 94.7 & 95.6 & 74.3 & 83.3 \\
\textsc{LaBSE} \citep{labse_acl22} & {93.9} & 95.5 & 94.4 & {97.5} & 95.3 & 92.1 & {83.3} & {95.4} & {96.1} & {75.4} & {85.9}\\
\midrule
\rowcolor{Gray}
\textsc{Ours} & \textbf{95.6} & \textbf{96.8 }& \textbf{97.3 }& \textbf{98.1} & \textbf{96.3 }& \textbf{94.3 }& \textbf{85.2 }& \textbf{96.5 }& \textbf{96.3} & \textbf{77.2 }& \textbf{86.3} \\
\bottomrule

\end{tabular}
\label{tb:tatoeba_details}

\end{table*}

\begin{table*}[!t]
\centering
\small
\caption{The effect of $\lambda$ on business corpora with Recall@30.}
\begin{tabular}{l|lll|llll|llll|l}

\toprule
 \multirow{2}{*}{$\lambda$} &\multicolumn{3}{c|}{AliExpress} &\multicolumn{4}{c|}{LAZADA} &\multicolumn{4}{c|}{DARAZ} &\multirow{2}{*}{Avg.}   \\\cmidrule{2-12}
                          & Ar   & En  & Zh  & Id   & Ms   & Fil   & Th   & Ur & Bn & Ne & Si &  \\
\midrule

$0.1$ & 87.1 & 82.2 & 91.1 & 68.3  & 60.8 & 77.7 & 80.2 &  88.0  &  90.4 & 60.2 & 84.0 &  \cellcolor{Gray}79.1 \\
\rowcolor{Gray}
$0.2$ & 85.4 & 81.9 & 91.1 & 70.9  & 58.8 & 77.6 & 82.2 &  87.9  &  88.4 & 65.2 & 81.6 & \textbf{79.2}  \\
$0.3$ & 81.1 & 81.7 & 90.6 & 70.7 & 59.3 & 78.1 & 76.8  &  88.2  &  89.8 & 62.0 & 83.6 & \cellcolor{Gray}78.3 \\
$0.4$ & 82.9 & 81.2 & 91.7 & 70.9 & 58.8 & 74.3 &  82.3 &  87.6 &   88.6 & 63.0 & 83.8 & \cellcolor{Gray}78.6 \\
$0.5$ & 87.2 & 80.2 & 90.6 & 70.8 & 60.6 & 72.8 &  82.2 &  88.1 &   89.2 & 64.9 & 82.3 & \cellcolor{Gray}79.0 \\
\bottomrule

\end{tabular}

\label{tb:dif_lmda}

\end{table*}

\begin{table*}[!t]
\centering
\small
\caption{The effect of code-switching rate (``$Cmd_r$") on business corpora with Recall@30.}
\begin{tabular}{l|lll|llll|llll|l}

\toprule
 \multirow{2}{*}{$Cmd_r$} &\multicolumn{3}{c|}{AliExpress} &\multicolumn{4}{c|}{LAZADA} &\multicolumn{4}{c|}{DARAZ} &\multirow{2}{*}{Avg.}  \\\cmidrule{2-12}
                          & Ar   & En  & Zh  & Id   & Ms   & Fil   & Th   & Ur & Bn & Ne & Si &  \\
\midrule
$0\%$ & 86.7 & 82.2 & 89.9 & 67.5 & 58.6 & 75.4 & 77.4 & 82.7 & 87.9 & 57.3 & 76.8 & \cellcolor{Gray}76.6  \\
\rowcolor{Gray}
$10\%$  & 89.0 & 82.2 & 92.4 &  73.7 & 61.3  & 78.4  &  80.2  &  84.6  &  90.8  & 60.2 & 79.0 & \textbf{79.3} \\
$20\%$  & 87.1 & 82.2 & 91.4 &  68.3 &  60.8 &  77.4 &  80.2  &  88.0  &  90.4  & 60.2 & 84.0 & \cellcolor{Gray}79.1  \\
$30\%$  & 84.5 & 82.2 & 92.4 &  70.5 &  59.2 &  76.6 &  81.7  &  87.7  &  89.3  & 66.1 & 74.4 & \cellcolor{Gray}78.6 \\
$40\%$  & 86.2 & 82.2 & 92.3 &  69.8 &  61.3 &  75.8 &   79.8 &  82.3  &  84.2  & 65.2 & 78.0 & \cellcolor{Gray}77.9 \\
$50\%$  & 85.1 & 82.2 & 91.9 &  69.7 &  59.3 &  76.4 &   81.1 &  85.9  &  79.9  & 65.5 & 78.4 & \cellcolor{Gray}77.8 \\
\bottomrule

\end{tabular}

\label{tb:dif_cmd_rate}

\end{table*}

\section{Results on Business Datasets}

As shown both in Table \ref{tb:XLSR_business_data_top10} and Table \ref{tb:XLSR_business_data_top20}, we make some explorations for the retrieving skill of our introduced approach on Ali-Express, DARAZ, and LAZADA corpora by evaluating Recall@\{10,20\}, respectively.
The conduction of our experiment is composed of two steps: first, we continually pre-train the models by combining the queries and labels among the training set. Then we conduct the fine-tuning on different languages by exploiting their own train set $\&$ dev set. 
Similar to the results on TOP-30 queries (see Table \ref{tb:XLSR_business_data}), VECO also obtains higher results among the baselines systems. However, our proposed model achieves remarkably better results than all baselines on each of the languages of the three business datasets.

\section{Results on the Task of STS}

We regard that there exists a bit of difference between the two tasks, such as semantic retrieval (SR) and semantic textual similarity (STS). But both of them take sentence-level representation as a backbone.
Therefore, we make some investigations on the task of STS.
As shown in Table \ref{tb:sts_task_result}, we also make further validation on the similar task STS by comparing the semantic representation skill of the baseline models and our proposed method. 
In this experiment, all of the test sets (STS12, STS13, STS14, STS15, STS16, STS-B, and SICK-R) only include English sentences. Thus we evaluate the baselines and our model only using the continually pre-trained model instead of the fine-tuned model.
We leverage Spearman's rank correlation coefficient to measure the quality of all models.
Among the baseline systems, the \textsc{SimCSE-BERT}$_{Large}$ obtain higher results than other baseline approaches, but our presented method steadily outperforms all the baseline models.

\section{Results on Publicly Open Corpora}

We conduct meaningful experiments on well-known and broadly used open available public datasets AskUbuntu and Tatoeba benchmark for sentence-level semantic retrieval.
In contrast with the experiment on the AskUbuntu dataset,
in this experiment, we continually pre-train all the models by leveraging the BUCC2018 corpus for the continual pre-training step. 
Due to the BUCC2018 corpus containing the train set and dev set, for a fair comparison, we also fine-tune our continually pre-trained model via the BUCC2018.
Moreover, the Tatoeba dataset covers more than 40 languages (shown with their ISO 639-1 code for brevity). But in our experiment, we only choose the part that differs from our business corpora, such as Afrikaans (af), German (de), Spanish (es), French (fr), Italian (it), Japanese (ja), Kazakh (kk), Dutch (nl), Portuguese (pt), Swahili (sw) and Telugu(te). 
As shown in Table \ref{tb:tatoeba_details}, all the detailed reviews about the comparison results are evaluated by accuracy on each language for Tatoeba. In this experiment, we choose 11 languages.

\section{The Effect of the Hyper-parameter}

As shown in Table \ref{tb:dif_lmda}, we make further validation on both business data and publicly available open corpora with different values of $\lambda$. 
For selecting the values of $\lambda$, we fix the code-switching rate $Cmd_r$ during the continual pre-training stage for our model.
The experimental result shows when the $\lambda = 0.2$, our model achieves better results than other values. Thus, we set $0.2$ as the default value of $\lambda$ 
in all experiments.

Additionally, as given in Table \ref{tb:dif_cmd_rate}, we also investigate 
the different values of the code-switching rate $Cmd_r$.
Similarly, we fix the lambda $\lambda$ during the continual pre-training step to select the proper values of $Cmd_r$ for our model.
It is not hard to infer from the experimental results that, when the $Cmd_r = 10\%$, our proposed approach obtains the highest performance compared with other values of $Cmd_r$.
Therefore, we take the  $10\%$ as a default value for $Cmd_r$.
The values of $\lambda$ and $Cmd_r$ are identical for the business and open corpora in the whole experiment.

\end{CJK}
\end{document}